\documentclass[runningheads]{llncs}

\usepackage{eccv}

\usepackage{eccvabbrv}
\usepackage{graphicx}
\usepackage{booktabs}
\usepackage[accsupp]{axessibility}  

\usepackage{multirow}
\usepackage{multicol}
\usepackage{booktabs}
\usepackage{graphicx}
\usepackage[export]{adjustbox}
\usepackage{pifont}
\usepackage{amssymb}
\usepackage{amstext}
\usepackage{amsmath}
\usepackage[bb=boondox]{mathalfa}
\newcommand{\cmark}{\ding{51}}%
\newcommand{\xmark}{\ding{55}}%
\usepackage{float}
\usepackage{graphicx}
\usepackage{amsmath}
\usepackage{amssymb}
\usepackage{booktabs}
\usepackage{siunitx}
\usepackage{bm}
\usepackage{subcaption}
\usepackage{soul}
\usepackage[dvipsnames]{xcolor}
\usepackage{array}
\usepackage{pifont}
\usepackage[accsupp]{axessibility}

\newcolumntype{P}[1]{>{\centering\arraybackslash}p{#1}}

\definecolor{orange}{RGB}{255,127,0}
\definecolor{green}{HTML}{82B366}
\definecolor{red}{HTML}{B85450}
\definecolor{red2}{HTML}{B85550}
\definecolor{unified}{HTML}{01994D}
\definecolor{disjoint}{HTML}{999900}

\newcommand{\naq}{\textbf{\texttt{\small NaQ}} }

\usepackage[pagebackref,breaklinks,colorlinks,citecolor=eccvblue]{hyperref}
\usepackage{hyperref}
\usepackage[capitalize]{cleveref}

\usepackage{orcidlink}
\newcommand{\name}{RGNet}
\newcommand{\quotes}[1]{``#1''}
\definecolor{orange}{RGB}{255,127,0}
\definecolor{green}{HTML}{82B366}
\definecolor{red}{HTML}{B85450}
\definecolor{red2}{HTML}{B85550}
\definecolor{unified}{HTML}{01994D}
\definecolor{disjoint}{HTML}{999900}

\begin{document}

\title{RGNet: A Unified Clip Retrieval and Grounding Network for Long Videos} 

\titlerunning{RGNet}

\author{
Tanveer Hannan$^{1,2}\thanks{Corresponding author: hannan@dbs.ifi.lmu.de}$ \quad
Md Mohaiminul Islam$^{3}$ \quad
Thomas Seidl$^{1,2}$ \quad
Gedas Bertasius$^3$\\ 
}

\authorrunning{Hannan et al.}

\institute{LMU Munich  \and
MCML  \and
UNC Chapel Hill}

\maketitle
\begin{abstract}
Locating specific moments within long videos (20–120 minutes) presents a significant challenge, akin to finding a needle in a haystack. Adapting existing short video (5–30 seconds) grounding methods to this problem yields poor performance. Since most real-life videos, such as those on YouTube and AR/VR, are lengthy, addressing this issue is crucial. Existing methods typically operate in two stages: clip retrieval and grounding. However, this disjoint process limits the retrieval module's fine-grained event understanding, crucial for specific moment detection. We propose RGNet which deeply integrates clip retrieval and grounding into a single network capable of processing long videos into multiple granular levels, e.g., clips and frames. Its core component is a novel transformer encoder, RG-Encoder, that unifies the two stages through shared features and mutual optimization. The encoder incorporates a sparse attention mechanism and an attention loss to model both granularity jointly. Moreover, we introduce a contrastive clip sampling technique to mimic the long video paradigm closely during training. RGNet surpasses prior methods, showcasing state-of-the-art performance on long video temporal grounding (LVTG) datasets MAD and Ego4D. The code is released at \href{https://github.com/Tanveer81/RGNet}{https://github.com/Tanveer81/RGNet}.
\keywords{Long Video Temporal Grounding \and Moment Localization}
\end{abstract}
    
\section{Introduction}
\label{sec:intro}
\begin{figure}[!t]
    \centering
    \includegraphics[width=.75\linewidth]{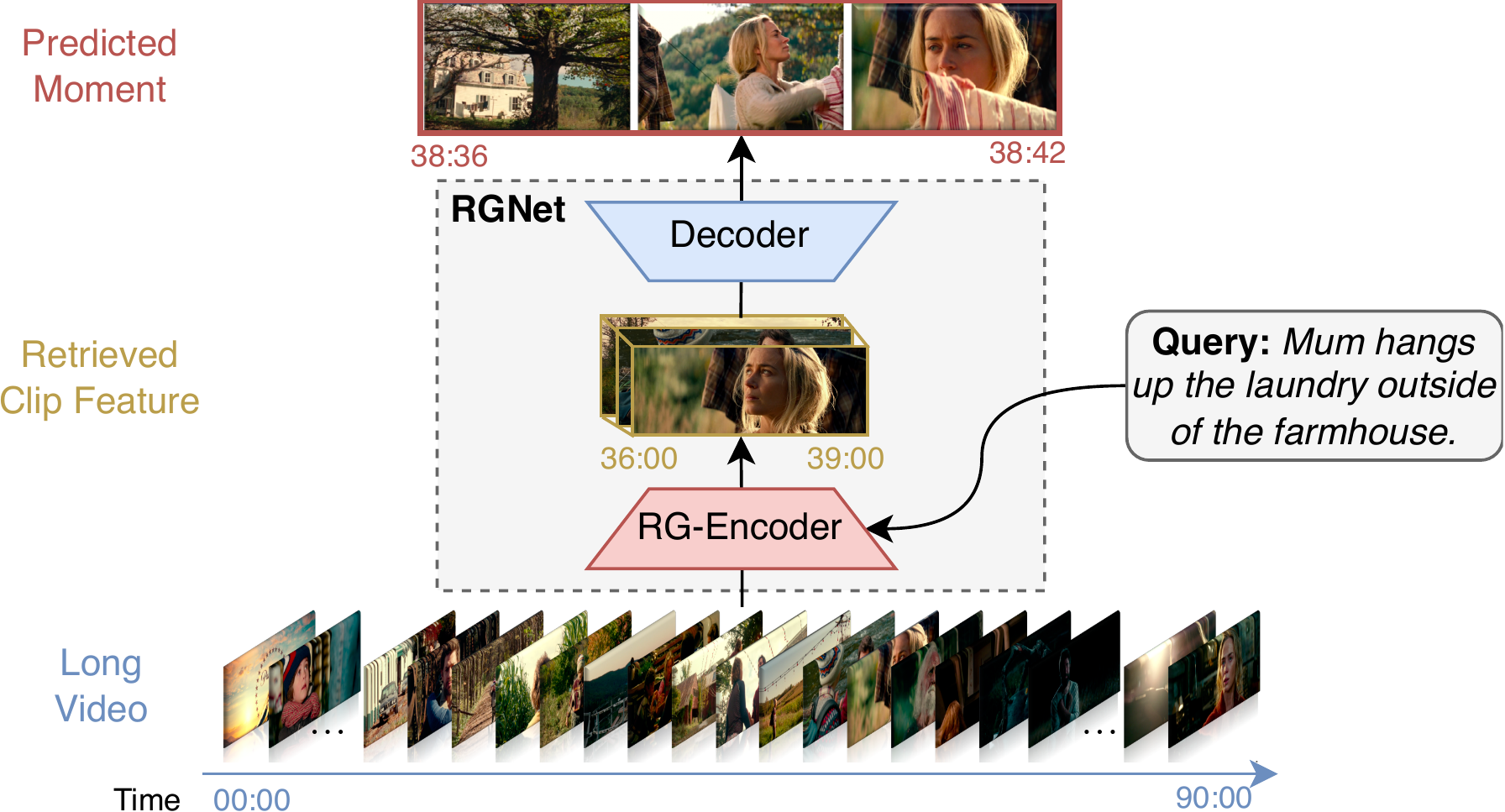}
    \caption{\textbf{Overview of RGNet.} It predicts the moment boundary specified by textual queries from an hour-long video. First, our proposed \textbf{RG-Encoder} maps the video and text features to a joint space and retrieves the relevant clip feature. The subsequent grounding decoder processes the retrieved features to predict the beginning and end times of the moment. The encoder parallelly operates at multiple levels of granularity (e.g., clip and frame) to achieve an end-to-end-solution.}
    \vspace{-2em}
    \label{fig:rgnet}
\end{figure}

The exponential rise in online videos has created a demand for effective content retrieval systems. The retrieval results of user-defined queries, particularly in long videos (20-120 minutes), often require locating specific moments. Most existing solutions\cite{zhang2020learning, zhang2020span, soldan2021vlg, chen2023joint}, tailored for short videos (5-30 seconds), struggle when applied to hour-long videos, slowing down the retrieval and impacting the quality of the results \cite{soldan2022mad, zhang2022elements}. This challenge emerges because it is very difficult to locate short moments in a long video based on text queries, a task known as Long Video Temporal Grounding (LVTG) \cite{grauman2022Ego4D, soldan2022mad}.

A straightforward solution \cite{hou2022cone,pan2023scanning} for the LVTG task is to divide the video into shorter clips, retrieve the most relevant one, and apply a grounding network to predict the moment. However, the grounding network cannot rectify failure in the clip retrieval phase. 

Empirically, we evaluate the impact of the two stages in Sec. \ref{sec:motivation} and identify clip retrieval as the primary factor for poor performance. 

\begin{figure}[!t]
    \centering
    \includegraphics[width=.75\linewidth]{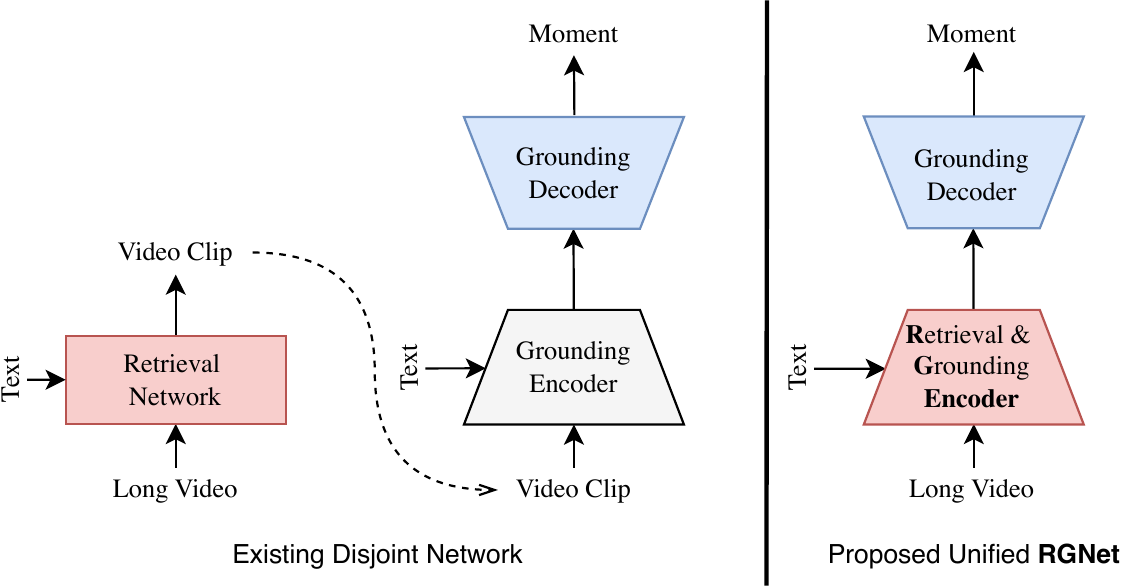}
    \caption{\textbf{Unified Solution. (left)} Existing methods involve a separate retrieval and grounding network. The disjoint retrieval lacks fine-grained event understanding, which is crucial for moment localization. \textbf{(right)} Our unified network architecture overcomes it by deeply integrating the retrieval module with the grounding objective.}
    \vspace{-2em}
    \label{fig:main} 
\end{figure}

The existing methods have a separate module for clip retrieval, which is disjoint from their grounding network (see Fig. \ref{fig:main}). They typically follow a text-video retrieval\cite{gorti2022x, lei2021less, liu2022ts2} technique to select the relevant clip. However, video retrieval only requires a high-level understanding of video topics. For example, a typical video retrieval query is \quotes{A movie about a family surviving in a farmhouse.}. In contrast, an example clip retrieval query in Fig. \ref{fig:rgnet} is \quotes{Find the moment when the mum hangs up laundry outside the farmhouse}. These specific moments from a long video require fine-grained event understanding. Thus, video retrieval models are suboptimal for these fine-grained moment localization tasks. 

We attribute this disjoint retrieval to the poor performance of these models. However, unifying clip selection and grounding is challenging due to their distinct setups. The former is a retrieval task, whereas the latter is a regression task. Moreover, it requires modeling two levels of video granularity: clip and frame. To address these technical challenges, we propose \textbf{RGNet}, a unified clip {R}etrieval and {G}rounding {Net}work (see Fig. \ref{fig:rgnet}). The single network enables end-to-end training of both stages. This improves the retrieval module's {fine-grained event understanding} by directly optimizing it with the moment annotations. Parallelly, the grounding network also benefits from the strong multimodal features produced by our encoder, further enhancing the localization capability. 

To achieve this unification, we propose a novel transformer encoder, {RG-Encoder}, which enables the grounding network to perform clip retrieval. This eliminates the suboptimal video retrieval network and effectively models long video clip retrieval. To enable the retrieval module to understand fine-grained events, we design a sparse-attention mechanism in the encoder. The sparse attention enables the retrieval module to focus on specific events in the video, which is more synergistic with the grounding task. Thus RG-Encoder is capable of operating both at clip and frame level granularity. We propose an Intra-Clip Attention Loss, which motivates the sparse attention to focus more on the video frames aligned with the specified event. Furthermore, we propose a negative clip mining technique to simulate clip retrieval from long videos during training. This negative mining enables our network to train on a large batch size with an inter-clip contrastive loss. The large number of negative clips in the batch closely mimics the long video paradigm and reduces the gap between the training and test phases.
 
Together, these components of \name{} bridge the gap between the two stages of LVTG, enhancing the fine-grained event understanding in hour-long videos. In summary, our contributions are fourfold:
\vspace{-.5em}
\begin{itemize}
    \item We systematically deconstruct existing LVTG methods into clip retrieval and grounding stages. Through empirical evaluations, we discern that disjoint retrieval is the primary factor contributing to poor performance.
    \item Based on our observations, we introduce RGNet, which integrates clip retrieval with grounding through parallel clip and frame-level modeling. This obviates the necessity for a separate video retrieval network, replaced instead by an end-to-end clip retrieval module tailored specifically for long videos.
    \item We introduce sparse attention to the retriever and a corresponding loss to model fine-grained event understanding in long-range video. We propose a contrastive negative clip-mining strategy to simulate clip retrieval from a long video during training. 
    \item  RGNet achieves state-of-the-art performance across both LVTG datasets, Ego4D \cite{grauman2022Ego4D} and MAD \cite{soldan2022mad}. For instance, RGNet outperforms the previous best Ego4D method~\cite{ramakrishnan2023naq} by a substantial margin (\textbf{9.7\%}). 
\end{itemize}

\section{Related Literature}
\label{sec:rel_lit}
\textbf{Short Video Temporal Grounding.} 
The recent grounding methods mainly focus on short videos. The best-performing ones utilize transformer variants \cite{carion2020end, zhu2020deformable, conditionaldetr, dabdetr}. For example, Moment-DETR \cite{lei2021detecting} adopted transformers for combined video and text features. Subsequent works, such as UMT \cite{liu2022umt} and QD-DETR \cite{moon2023query}, split the cross-modal and unimodal modeling of video and text features for enhanced performance. EATR \cite{jang2023knowing} improves its decoder with video-specific moment queries. These models were initially designed for shorter videos. However, when applied to hour-long videos, the moments become extremely tiny, posing a challenge akin to finding a needle in a haystack. Hence, these methods perform poorly on long videos, as noted by the authors of the MAD dataset \cite{soldan2022mad}. In contrast, we effectively address the task by simultaneously processing long videos at two granular levels (video clips and frames) within a single network.

\hspace{1pt}\\
\textbf{Long Video Temporal Grounding.} Recently, video temporal grounding has been adapted for long videos with MAD \cite{soldan2022mad} and Ego4D \cite{grauman2022Ego4D} datasets. Typically, the methods designed for long videos involve two stages. Proposal-free methods \cite{zhang2020learning,zhang2020span,soldan2021vlg, barrios2023localizing, liu2022reler} segment lengthy videos into smaller parts to predict candidate moments and then rank them to obtain the final predictions. Proposal-based methods \cite{pan2023scanning,hou2022cone} generate proposal clips or anchors and subsequently apply their grounding model to the retrieved proposals. Some methods, like CONE \cite{hou2022cone}, and M-Guidance \cite{barrios2023localizing}, utilize the detection transformer~\cite{carion2020end} as the grounding network to improve the grounding phase. These proposal-based models are more efficient and perform better than their counterparts. However, they need a separate retrieval module to select the proposal clips. This disjoint two-stage architecture is what we find to be the main reason for the poor performance of these models. In contrast, we propose a novel transformer encoder that solves both clip retrieval and grounding in a unified way, enabling end-to-end optimization.

\hspace{1pt}\\
\textbf{Video-text Retrieval.}
The early retrieval methods ~\cite{chen2020fine,dong2021dual,kiros2014unifying,yu2018joint,yu2017end} designed multimodal fusion techniques to align pre-trained video and text models. To bridge the gap between pre-training and downstream tasks, large-scale pre-trained video-language representations \cite{bain2021frozen,gabeur2020multi,ge2022bridging,VindLU_CVPR2023, luo2020univl,wang2022all,xu2021vlm,xu2021videoclip,xue2022advancing} have been introduced in various works. Several studies~\cite{fang2023uatvr,gao2021clip2tv,jiang2022cross,jin2022expectation,ma2022x,wang2022disentangled,xue2022clip,zhang2023multimodal, wang2023unified} have transferred the knowledge from CLIP~\cite{radford2021learning} to text-video retrieval tasks. In recent studies, various sampling strategies \cite{lei2021less, gorti2022x, liu2022ts2}  have been explored, enabling the model to selectively concentrate on pertinent video frames based on provided text inputs. For instance, Clip-BERT \cite{lei2021less} introduces a sparse sampling strategy, X-Pool~\cite{gorti2022x} utilizes a cross-modal attention model, and TS2-Net~\cite{liu2022ts2} adapts CLIP to capture temporal information and eliminate unimportant tokens. While existing methods aim to retrieve high-level video topics from a pool of video collections, moment localization demands a fine grained understanding to identify specific events within a video. Despite the similar setup to video retrieval, retrieving the corresponding clip from a long video necessitates deeper event comprehension. Therefore, we tailored our encoder to enhance event understanding and simulated precise clip retrieval conditions during training.

\section{RGNet}
\label{sec:meth}
\begin{figure}[!h]
    \centering
    \includegraphics[width=.5\linewidth]{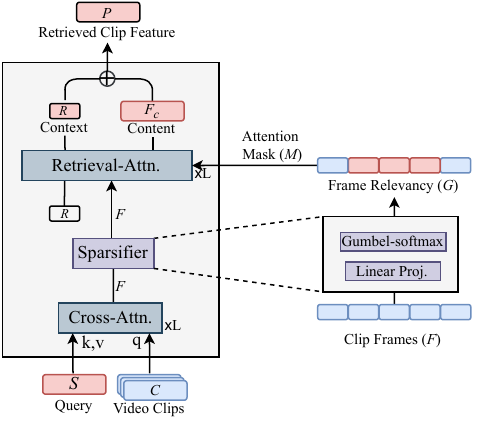}
    \caption{\textbf{Overview of RG-Encoder.} It takes video clips and textual query as input and retrieves the relevant clip features. 
    First, a cross-attention fuses the clips with text, and the sparsifier masks the out-of-moment frames. Based on the mask, the retrieval attention focuses on in-moment frames (colored \textcolor{red}{red}) and generates clip-level context and frame-level content features. We combine the context and content to generate the retrieved clip feature. }
    \vspace{-3em}
    \label{fig:rg-encoder}
\end{figure}

This section presents a detailed description of our proposed unified LVTG method. As illustrated in Fig. \ref{fig:rgnet}, RGNet takes a long video and a text query as input and predicts the precise moment semantically correlated with the query in an end-to-end manner. The whole network is divided into the proposed RG-Encoder (Fig. \ref{fig:rg-encoder}) and a decoder (Sec. \ref{decoder}). The encoder retrieves the most relevant clip feature from the input long video. The decoder then processes the retrieved clip feature to predict the precise moment boundary. Additionally, an Intra-Clip Attention Loss and an Inter-Clip Contrastive Loss are incorporated to facilitate low-level event understanding in long-range videos.
\subsection{Feature Extraction}
Let us assume we have an untrimmed long video $V$ consisting of $T$ frames and a query sentence $S$ with $N$ words corresponding to a target moment's center $\tau_{c}$ and width $\tau_{w}$. Following previous methods \cite{hou2022cone,pan2023scanning}, we use pre-trained frozen models to extract visual features ${V}=\{{f}^{1}, {f}^{2}, ..., {f}^{T}\}\in{R}^{T \times D_f}$ as well as textual features ${S}=\{{w}^{1}, {w}^{2}, ..., {w}^{N}\}\in{R}^{N \times D_w}$, where $D_f$ and $D_w$ represent the feature dimensions of video frames and query words. We slice the long video $V$ into clips of length $L_c$ and feed them into our model. We employ a sliding window of length $L_c$  and stride $L_{c}/2$ to obtain a collection of $T_c$ clips ${C}=\{{C}^{1}, {C}^{2}, ..., {C}^{T_c}\}\in{R}^{T_c \times L_{c} \times D_f}$. Here, the $i^{th}$ clip, ${C}^i=\{{f}^{s_i+1},{f}^{s_i+2},...,{f}^{s_i+L_c}\} \in {R}^{L_c \times D_f}$, where $s_i$ is the start index of the clip.

\subsection{RG-Encoder}
All the clips ${C}$ from a single video comprise samples in a batch. The RG-Encoder (Fig. \ref{fig:rg-encoder}) processes them with the query ${S}$ to retrieve the relevant clip feature $P$. First, the cross-attention generates query-conditioned frame features ${F}$. Then, based on the frame-query correlation calculated from $F$, the sparsifier determines the relevant frames inside the clip to produce a mask ${M}$ for retrieval attention. We incorporate a learnable token ${R}$ to capture the context of the clip. Additionally, the frame features ${F}$ undergo retrieval attention based on the predicted mask $M$ to generate the content feature $F_c$. The combination of context $R$ and content $F_c$ forms the retrieved clip feature $P$. 

\hspace{1pt} \\
\textbf{Cross Attention:}
To evaluate the relevance of each frame within the clip $C^i$, we utilize cross-attention, where frame features act as queries and text features serve as keys and values. Cross-attention represents the frame features as a weighted average of the textual features, scaled by their mutual correlation. This process results in text-conditioned frame features $F^i$, establishing a fine-grained correspondence between the modalities. The query, key and values are calculated as ${Q}^{i} = l_Q({C}^{i}) \in {R}^{L_c \times D_f}$, ${K} = l_K({S}) \in {R}^{N \times D_q}$ and ${V} = l_V({S}) \in {R}^{N \times D_q}$ respectively. Here, $l_Q(\cdot)$, $l_K(\cdot)$, and $l_V(\cdot)$ denote the projection layers for the query, key, and value.
Then, we apply the cross-attention layer as follows:
\begin{equation}
\begin{gathered}
    F^{i} = \operatorname{softmax}({{Q}}^i{K}^T){V} +  {Q}^{i} \\
\end{gathered}
\end{equation}
\hspace{1pt} \\
\textbf{Sparsifier:}
To retrieve the clip, we need to assess the relevance of each clip based on the text query. Clip relevance is determined by aggregating the relevancy of its frames. However, given the small moment duration, most frames are unrelated to the query. Hence, we classify frames into relevant and non-relevant categories using the sparsifier to assist retrieval attention in focusing on these fine-grained events.
We calculate the relevancy ${G}^{j} \in [0,1]$ of the ${j}^{th}$ frame with Eq. \ref{eqn:sampler}. We use a differentiable Gumbel-softmax function \cite{hannan2023gratt, gumble-softmax} for enabling end-to-end training. The detailed calculation is in the supplementary materials.
\begin{equation}
{G}^{j} = \operatorname{Gumbel-softmax}(\text{linear}({f}^{j})) 
\label{eqn:sampler}
\end{equation}

The intra-clip attention loss optimizes this classification using the ground truth frame relevancy information described in Sec. \ref{loss}. To distinguish irrelevant frames from their counterparts, we constrain their updates in the retrieval attention. This restriction is implemented with an attention mask computed from the learned frame relevancy ${G}^{j}$. The attention mask ${M}$ between $j^{{th}}$ and $k^{{th}}$ frame is calculated with Eq. \ref{eq:mask}. Here, the $j^{th}$ and $k^{th}$ frames are the attention query and key, respectively. This leads to a reduced set of relevant and informative frames for subsequent attention.

\begin{equation}
\label{eq:mask}
   {M}(j, k)=\left\{\begin{array}{ll}
0 & \text { if } {G}^{j}>0.5 \hspace{5pt} \text{or} \hspace{5pt} j = k\\
-\infty & \text { otherwise } 
\end{array} \right. 
\end{equation}
\hspace{1pt} \\
\textbf{Retrieval Attention:}
We aim for a learnable clip retrieval function to train end-to-end with the grounding objective. Thus, we devise the retrieval attention for aggregating the text-conditioned frame features based on their relevancy. To condense the $i^{th}$ clip into a contextual feature embedding, we introduce a randomly initialized learnable vector ${R}^i$. We denote this as retrieval token, which is trained with the inter-clip contrastive loss. 

First, we concatenate the retrieval token with the text-conditioned frame features to create the queries ${Q}^{i}=[{R}^i,{F}^{i}]$ for retrieval attention. We also generate an attention mask containing all zeros for the retrieval token and concatenate it with ${M}$. We set the {key} ${K} = {Q}^{i}$, and the {value} ${V} = {Q}^{i}$ to calculate the retrieval attention according to Eq. \ref{eq:ret_attn}. We get the clip-level context feature, ${R}^i = \Tilde{Q}^{i,1}$ and frame-level content features, ${F_c}^{i,j}=[\Tilde{Q}^{i,2},\Tilde{Q}^{i,3},...,\Tilde{Q}^{i,L_c+1}]$. Here, ${F_c}^{i,j}$ is the $j^{th}$ frame feature of the $i^{th}$ clip. 
\begin{equation}
\begin{gathered}
    {\Tilde{Q}}^{i} = \operatorname{softmax}({Q}^{i}{K}^{T}+ {M}){V} +  {Q}^{i} \\
\end{gathered}
\label{eq:ret_attn}
\end{equation}

Further, the context feature R is utilized to retrieve the clip. It undergoes a linear projection layer, $l_s(\cdot)$, to obtain context scores ${S_r} = \text{sigmoid}(l_s({R}))$. We retrieve the most relevant clip based on the context score ${S_r}$. Importantly, the context R represents the queried moment by attending to the reduced set of relevant frames. So, we fuse it with the frame features with Eq. \ref{eq:fuse} to produce stronger multimodal inputs for the decoder. It enhances the decoder's localization capability significantly. The decoder only processes features from the retrieved clip to predict the moment.
 \begin{equation}
{P}^{i,j} = {F_c}^{i,j} + {R}^i \times {G}^{j}, \hspace{1pt} i \in P, \forall j\\
\label{eq:fuse}
\end{equation} 

\subsection{Grounding Decoder}
\label{decoder}
\vspace{-.5em}
We process the retrieved clip features with a transformer decoder to predict precise moment boundaries. Previous methods \cite{lei2021detecting, hou2022cone} feed both modalities into the decoder, introducing multiple positional information (e.g., time in the video and word order in the text) that complicates the detection task. In contrast, our encoder injects text information directly into the clip, eliminating the requirement to feed both modalities into the decoder. We incorporate learnable anchor queries \cite{dabdetr} to represent the moment's center and width as $(\tau_{c}, \tau_{w})$. The decoder has 2 cross-attention and 2 self-attention layers.

\subsection{Loss Functions} 
\label{loss}
\textbf{Intra-Clip Attention Loss} guides our sparsifier to distinguish frames within and outside of the ground truth moment. This enables the retrieval attention to focus on relevant video regions and achieve fine-grained event understanding. Specifically, the in-moment frames are required to maintain higher relevancy scores than those outside the ground truth moment. The loss is defined as:
\begin{align}
\begin{split}
\mathcal{L}_{\text{attn}} = & \operatorname{max}(0, \Delta + S_c(i,j_{\text{out}}) - S_c(i,j_{\text{in}}))
\end{split}
\end{align}
In the equation, $S_c$ is the relevancy score, $\Delta$ represents the margin, and anchor $i$ is an in-moment frame. We randomly chose in-moment and out-of-moment frames $j_{\text{in}}$ and $j_{\text{out}}$ from the same clip. To calculate the relevancy score, the context features ${R_i}$, and encoded clip $P^{i,j}$ undergoes linear layers , $l_{r}(\cdot)$ and $l_{c}(\cdot)$, to obtain projected context ${R}^i_{{proj}} = l_{r}({R^i}) \in {R}^{D_l}$ and content features, $P^{i,j}_{proj} = l_{c}(P^{i,j}) \in {R}^{{L_c}\times{D_l}}$. Then, we calculate the frame-level relevancy score: 
\begin{equation}
     S_c(i,j) = {R}^i_{{proj}} \cdot P^{i,j}_{{proj}}
\label{eq:samp_loss}
\end{equation}

\hspace{1pt} \\
\textbf{Inter-Clip Contrastive loss.} 
A long video contains many clips with similar environments and scenes. However, most of these clips do not contain the specific moment we are searching for. So, high-level video understanding is not enough to distinguish precise moments. During the test phase, we need to handle many such negative clips. To tackle this issue, unlike previous approaches~\cite{hou2022cone}, we sample a large number of negative clips and train our retriever on a large batch size. Hence, this negative clip sampling reduces the disparity between the train and test phases.

For this section, we denote the context feature, $R^{i,j}$ for $i^{th}$ proposal and $j^{th}$ text query. We train our model contrastively with positive text-clip pairs $[R^{i,j}]_{i = j}$ and negative pairs $[R^{i,j}]_{i \neq j}$. We use the InfoNCE loss \cite{oord2018representation} to identify the positive text-clip pair amongst a set of unrelated negative samples. Here, $l_{cont}(\cdot)$ is a linear projection layer that transforms the $D_f$ dimensional context feature into a one-dimensional logit.
\begin{equation}
    \mathcal{L}_{\text{cont}} = - \sum_{i}  log \frac{\exp(l_{\text{cont}}({R^{i,i}}))}{\sum_{j}\exp(l_{\text{cont}}({R^{i,j}}))} 
\label{equ: adapt}
\end{equation}

\hspace{1pt} \\
\textbf{Grounding loss.}
The objective functions for grounding, which aim to locate desired moments, are adopted from the baseline approach~\cite{lei2021detecting}. The grounding loss $\mathcal{L}_{\text{g}}$ measures the discrepancy between the ground truth (GT) and predicted moments. A one-to-one correspondence between GT and predicted moments is established using the Hungarian algorithm. This loss includes both an $\mathcal{L}1$ loss and a generalized IoU loss ($\mathcal{L}{\text{gIoU}}$). Additionally, a cross-entropy loss $\mathcal{L}_{\text{CE}}$ is employed to classify predicted moments as either foreground or background.
Thus, $\mathcal{L}_{\text{g}}$ is defined as follows:
\begin{eqnarray}
    \label{eqn moment localization loss}
    &\mathcal{L}_{\text{g}} = \lambda_{\mathcal{L}1} ||\tau - \hat{\tau}|| + \lambda_{\text{gIoU}} \mathcal{L}_{\text{gIoU}}(\tau, \hat{\tau}) + \lambda_{\text{CE}} \mathcal{L}_{\text{CE}},
\end{eqnarray}
where $\tau$ and $\hat{\tau}$ represent the ground-truth moment and its corresponding prediction, containing center coordinates $\tau_{c}$ and width $\tau_{w}$. The hyper-parameters $\lambda_{*}$ are used for balancing the losses. Finally, with the attention and contrastive loss, the total loss $\mathcal{L}_{\text{total}}$ is defined as follows:
\begin{eqnarray}
    \label{eqn moment localization loss}
    &\mathcal{L}_{\text{total}} = \lambda_{\text{attn}} \mathcal{L}_{\text{attn}} + \lambda_{\text{cont}} \mathcal{L}_{\text{cont}} + \mathcal{L}_{\text{g}} 
\end{eqnarray}

\section{Experimental Setup}
\label{sec:exp}  
\subsection{Datasets and Evaluation Metric} 
\textbf{MAD} \cite{soldan2022mad} is an extensive dataset comprising 1.2K hours of full-length movies and 384K natural language queries, each associated with specific moments in the videos. The videos are orders of magnitude longer than previous datasets, with an average duration of 110 minutes, while the specified text moments are on average 4.1 seconds. This small moment-to-video ratio poses a significant challenge for the grounding task.

\hspace{1pt} \\
\textbf{Ego4D-NLQ} \cite{grauman2022Ego4D} is a large-scale egocentric video data set with multiple challenges. Specifically, we use the episodic memory benchmark Ego4D-NLQ, which requires localizing where the answer to a natural language query can be seen. It contains around $13$ template questions, amounting to around 74K queries. The train, val, and test sets contain 11.3K, 3.9K, and 4.0K queries. The video length ranges from 8 to 20 minutes, with an average of 8.25 minutes, and the average moment duration is 8.3 seconds. This means the moments constitute only 2\% of the input video on average.

\hspace{1pt} \\
\textbf{Grounding Metric:}
For grounding, we follow previous methods \cite{soldan2022mad, hou2022cone}, and adopt the standard metric Recall@$k$ at IoU=$\theta$ (R$k_{\theta}$). This metric represents the percentage of testing samples with at least one grounding prediction whose intersection over union (IoU) with the ground truth (GT) is larger than $\theta$ among the top-$k$ predictions.

\hspace{1pt} \\
\textbf{Retrieval Metric:}
To assess our proposal retrieval stage independently of the grounding, we utilize the standard retrieval metric \cite{gorti2022x}, Recall at Rank $k$ (R@$k$). This metric calculates the percentage of GT moments present in the top-$k$ retrieved proposals.

\subsection{Implementation Details:}
We use pre-trained CLIP \cite{radford2021learning} and EgoVLP \cite{lin2022egocentric} models to extract video frames from MAD and Ego4D. Text features for both datasets are extracted using CLIP. The feature extractors are frozen, and their pre-trained weights remain unchanged during training. For Ego4D, we pre-train our model with NaQ annotations from previous work \cite{ramakrishnan2023naq}. The proposal length, $W_c$ is set to 180s for MAD and 48s for Ego4D, with only the top-30 and top-5 proposals retrieved, respectively. Training on Ego4D and MAD is conducted on four Nvidia-RTX-A6000 GPUs while fine-tuning on Ego4D is performed on a single GPU. Considering the longer proposals in MAD, we utilize a batch size of 32 for Ego4D and 8 for MAD. The number of moment queries is set to 5. For loss computation, we set hyperparameters as $\lambda_{L1}=10$, $\lambda_{\text{gIoU}}=1$, $\lambda_{\text{CE}}=4$, $\lambda_{\text{samp}}=1$, $\lambda_{\text{cont}}=10$, and $\Delta=0.2$. We use Xavier initialization \cite{glorot2010understanding} and employ AdamW \cite{loshchilov2017decoupled} with an initial learning rate of $1 \times 10^{-4}$. RGNet is trained for 35 and 200 epochs on MAD and Ego4D datasets. The initial learning rate is reduced by one order of magnitude at epochs 25 for MAD and 120 for Ego4D.

\section{Results and Analysis}
We first compare our performance with previous state-of-the-art models. Then, we report detailed ablation studies of our proposed method and visualize the qualitative results compared to the disjoint baseline.
\begin{table}[!h]
    \begin{adjustbox}{width=1\textwidth}
    \small
    \begin{tabular}{lccccccrcccccccc}
    \toprule
    \multirow{2}{*}{\textbf{Model}} & \multirow{2}{*}{\naq} & 
    \multicolumn{5}{c}{\textbf{Ego4D-NLQ} \cite{grauman2022Ego4D}}   &       & \multicolumn{7}{c}{\textbf{MAD} \cite{soldan2022mad}}&  \multirow{2}{*}{\textbf{Avg}}\\
    \cmidrule{3-7}\cmidrule{9-15} &&R1$_{.3}$ & R5$_{.3}$ &R1$_{.5}$ & R5$_{.5}$ & Avg &   &R1$_{.1}$ & R5$_{.1}$  & R1$_{.3}$ & R5$_{.3}$  & R1$_{.5}$ & R5$_{.5}$ & Avg \\
    \midrule
    2D\textcolor{gray}{-}TAN\cite{zhang2020learning} & \textcolor{gray}{\xmark}  & 5.04 & 12.89 &  2.02 & 5.88 & 6.46 && 3.22 & 11.90 & 2.52  & 9.25 &  1.58  & 5.69  & 5.69  & 6.07 \\
    UniVTG\cite{lin2023univtg}  & \textcolor{gray}{\xmark}  &      11.74     &     7.54      &    3.25      &     7.88   &7.60&&\textcolor{gray}{-}&\textcolor{gray}{-}&\textcolor{gray}{-}&\textcolor{gray}{-}&\textcolor{gray}{-}&\textcolor{gray}{-}&\textcolor{gray}{-}&\textcolor{gray}{-}  \\
    VSLNet\cite{zhang2020span} &  \cmark   &   {10.26}   & {19.01}     &{5.81}     &{12.67}  & 11.93 &&\textcolor{gray}{-}&\textcolor{gray}{-}&\textcolor{gray}{-}&\textcolor{gray}{-}&\textcolor{gray}{-}&\textcolor{gray}{-}&\textcolor{gray}{-}&\textcolor{gray}{-} \\ 
    VLG-Net~\cite{soldan2021vlg} & \textcolor{gray}{\xmark}  & \textcolor{gray}{-}&\textcolor{gray}{-}&\textcolor{gray}{-}&\textcolor{gray}{-}&\textcolor{gray}{-}&& 3.64  & 11.66   & 2.76  & 9.31   & 1.65  & 5.99  & 5.84 &\textcolor{gray}{-}\\
    M-DETR\cite{lei2021detecting} & \textcolor{gray}{\xmark}  & 8.23 & 23.23 & 5.01 & 13.37 &12.46&& 3.60  & 12.98 & 2.81 & 9.86  & 1.67 & 5.58  & 6.08  & 9.27 \\
    M-Guide~\cite{barrios2023localizing}  & \textcolor{gray}{\xmark}  & \textcolor{gray}{-}&\textcolor{gray}{-}&\textcolor{gray}{-}&\textcolor{gray}{-}&\textcolor{gray}{-}&&9.30  & 18.96  & 4.65  & 13.06  & 2.16  & 7.40  & 9.26 &\textcolor{gray}{-}   \\
    SOONet\cite{pan2023scanning} & \textcolor{gray}{\xmark}  & 8.00 & 22.40 & 3.76 & 11.09 &11.31&& {11.26}  & {23.21}  & {9.00}  & \textbf{19.64}  & {5.32}  & \textbf{13.14}  & {13.59} & 12.45 \\
    H-Hands\cite{zhang2023helping} & \textcolor{gray}{\xmark}  & 13.20 & 23.30 & 7.90 & 15.60 &15.00&&\textcolor{gray}{-}&\textcolor{gray}{-}&\textcolor{gray}{-}&\textcolor{gray}{-}&\textcolor{gray}{-}&\textcolor{gray}{-}&\textcolor{gray}{-}&\textcolor{gray}{-} \\
    CONE~\cite{hou2022cone}  & \textcolor{gray}{\xmark}  & {14.15} & {30.33} & {8.18}  & {18.02}&{17.67}&&8.90 & 20.51 & 6.87 & 16.11  & 4.10  & 9.59 & 11.01 & {14.34}\\
    EgoVLP  &  \cmark   &   {15.90}   & {26.38}     &{9.46}     &{17.80}  & 17.38 &&\textcolor{gray}{-}&\textcolor{gray}{-}&\textcolor{gray}{-}&\textcolor{gray}{-}&\textcolor{gray}{-}&\textcolor{gray}{-}&\textcolor{gray}{-}&\textcolor{gray}{-} \\
    ReLeR \cite{ramakrishnan2023naq} &  \cmark& {19.31} & 23.62 &{11.59} &15.75 &17.57&&\textcolor{gray}{-}&\textcolor{gray}{-}&\textcolor{gray}{-}&\textcolor{gray}{-}&\textcolor{gray}{-}&\textcolor{gray}{-}&\textcolor{gray}{-}&\textcolor{gray}{-}  \\
    \textbf{Ours} \texttt{\small {(Default)}} & \textcolor{gray}{\xmark}  & {18.28} & {34.02} & {12.04}  & {22.89} &{21.81}&& \textbf{12.43} & \textbf{25.12} & \textbf{9.48}  & {18.72}   & \textbf{5.61}  & {10.86} & \textbf{13.70} & {17.80} \\
    \textbf{Ours} &  \cmark & \textbf{20.63} & \textbf{41.67} &\textbf{12.47} &\textbf{25.08} &\textbf{24.96}&& \textbf{12.43} & \textbf{25.12} & \textbf{9.48}  & {18.72}   & \textbf{5.61}  & {10.86} & \textbf{13.70} & \textbf{19.33}\\
    \bottomrule
    \end{tabular}
    \end{adjustbox}
    \caption{\textbf{Main results on Ego4D-NLQ and MAD.} RGNet achieves state-of-the-art performance on both datasets. Our default network is trained without NaQ annotations \cite{ramakrishnan2023naq} on Ego4D. Even without NaQ annotations, our default network shows a larger improvement, underscoring the effectiveness of our solution with limited data.
     }
    \vspace{-3.5em}
\label{tab:sota}
\end{table}

\subsection{Result on the Ego4D-NLQ Dataset}
RGNet demonstrates state-of-the-art performance on all metrics for the Ego4D-NLQ benchmark. As shown in Table \ref{tab:sota}, RGNet improves its primary evaluation metric, the R$1_{.3}$ and R$5_{.3}$ score average by 9.7\%. Separately, we outperform the previous best methods on R$1_{.3}$ and R$5_{.3}$ by \textbf{1.32\%} and \textbf{18.1\%}, respectively. This improvement is consistent across other metrics as well. The unified clip and frame-level modeling in RGNet parallelly capture long- and short-range temporal dependencies, contributing to its superior performance. By modeling the entire video as a whole, RGNet learns better cross-modal alignment by contrasting visual scenarios from various events within the same video. Notably, compared to SooNet \cite{pan2023scanning} and CONE \cite{hou2022cone}, two prominent proposal-based methods, RGNet achieves an average score improvement of $13.7\%$ and $7.3\%$. This substantial improvement is attributed to our unified modeling of clip retrieval, whereas they separately employ a retrieval similarity search between the video and the text. 

\subsection{Result on the MAD Dataset}
We compare our model's performance on the MAD dataset in Tab. \ref{tab:sota}. RGNet achieves state-of-the-art performance on both the R$1_{.1}$ and R$5_{.1}$ scores by outperforming the previous best SooNet \cite{pan2023scanning} by \textbf{1.2\%} and \textbf{1.9\%}. Together with its competitive performance on other metrics, RGNet demonstrates the importance of end-to-end modeling of long videos. Specifically, our performance of R$1$ is superior at all IoU thresholds, which testifies to the effectiveness of the proposed retrieval method. Notably, prior models rely on heuristic dot product similarity search applied to CLIP \cite{radford2021learning} features for the retrieval phase. In contrast, our shared retrieval feature and its unified optimization with the grounding objective drastically enhance the first stage retrieval, as seen in Tab. \ref{tab:rg_ablation}. It further helps the network capture more discriminative events crucial for detecting extremely tiny moments in hour-long videos. 

\subsection{Disjoint vs. Unified Architecture} 
\label{sec:motivation}
Since the long video temporal grounding (LVTG) performance relies on the first stage of clip retrieval, it does not perfectly reflect the actual moment localization capability of its grounding network. To assess the standalone grounding performance, we designed an oracle experiment operating exclusively on the clip where the ground truth moment is present. Compared to the oracle grounding, the LVTG $R1_{.3}$ drops by 15.7\% and 23.9\% for Ego4D and MAD in Tab. \ref{tab:rg_ablation}. The poor grounding in LVTG emerges from the suboptimal retrieval accuracy of the disjoint network. Hence, retrieving clips where the ground truth moment exists is crucial, independent of the subsequent grounding accuracy. Regardless of the effectiveness of moment boundary detection, if the clip containing the moment boundary cannot be accurately retrieved. 
\begin{table*}[!h]
\centering
\begin{tabular}{llrccrrccrcc}
\toprule 
\multirow{2}[2]{*}{\textbf{Data}} & \multirow{2}[2]{*}{\textbf{Model}} && \multicolumn{2}{c}{\textbf{Clip Retrieval}} && \multicolumn{3}{c}{\textbf{Oracle Grounding}} && \multicolumn{2}{c}{\textbf{LVTG}}\\
\cmidrule{4-5}\cmidrule{7-9} \cmidrule{11-12}
& & &R@1 & R@5 &&& R1$_{.3}$ & R5$_{.3}$ && R1$_{.3}$ & R5$_{.3}$ \\
\toprule
  \multirow{2}[2]{*}{Ego4d} & {{Baseline}} && 31.71 & 64.63  &&& 29.84 & 54.01  && {14.15} & {30.33} \\
  & \textbf{Ours} && \textbf{42.08}  & \textbf{76.28} & &&$\mathbf{36.53}$ & $\mathbf{63.64}$ && \textbf{20.63} & \textbf{41.67} \\
 \midrule  
 \multirow{2}[2]{*}{MAD} & {{Baseline}} &&  12.41 & 24.50 &&& 29.49 & 53.02  && 6.87 & 16.11  \\
  & {\textbf{Ours}} &&  \textbf{25.01}  & \textbf{50.02}   &&&$\mathbf{33.42}$ & $\mathbf{63.43}$  && \textbf{9.48}  & \textbf{18.72}\\
\bottomrule 
\end{tabular}
\caption{\textbf{Empirical impact of two stages.} To asses standalone grounding capability, we run an oracle experiment on the clip where the ground truth moment is present. The grounding capability degrades in LVTG evaluation because of incorrect selection by the disjoint clip retrieval network. Our unified model improves retrieval significantly, which leads to more effective temporal grounding in long videos.
}
\vspace{-2.5em}
\label{tab:rg_ablation}
\end{table*}

Our unified approach consistently improves all retrieval metrics compared to the disjoint baseline. Notably, we achieve a staggering $10.4\%$ and $12.6\%$ improved R@1 score for Ego4D and MAD. This improvement in clip retrieval leads to state-of-the-art LVTG performance. Moreover, the grounding also improves because of the stronger clip features generated by our encoder. Compared to the disjoint baseline, the oracle R$1_{.3}$ scores improved by about $6.7\%$ and $4.0\%$ for Ego4D and MAD, respectively. Hence, the unified end-to-end architecture proves more effective at LVTG by mutually improving both stages.

\subsection{Ablation Studies}
This section reports detailed ablation studies on the proposed modules and loss functions. We also ablate with various clip lengths and retrieved clip numbers. Unless stated, the experiments are done with the \texttt{\small {Default}} network w/o NaQ annotations \cite{ramakrishnan2023naq} for the Ego4D-NLQ dataset and CONE \cite{hou2022cone} as the baseline. 
\begin{table}[!h]
    \centering
    \setlength{\tabcolsep}{1.5mm}
    \begin{tabular}{l|cccc}
    \toprule 
    \textbf{Model}  & R1$_{.3}$  & R5$_{.3}$ & R1$_{.5}$  & R5$_{.5}$  \\
    \midrule
    {\textbf{RGNet}} \texttt{\small {(Default)}} & 
    \textbf{18.28} &  \textbf{34.02} & \textbf{12.04}    & \textbf{22.89} \\
    \midrule
    \hspace{3mm} w/o Retrieval Token & 17.80  &  33.99 &  11.25    &  22.32       \\
    \hspace{6mm} w/o Sparsifier & 16.12   &  31.57 &   9.91      &  20.47     \\
    \hspace{9mm} {w/o RG-Encoder$^{\ast}$} &  14.15  & 30.33 & 8.18  & 18.02    \\
    \midrule
    \hspace{3mm} w/o Contrastive Loss & 17.41   &  32.12 &   10.79      &  22.80        \\
    \hspace{6mm} w/o Attention Loss & 16.21   &  31.59 &   9.91      &  20.53     \\
    \bottomrule 
    \end{tabular}
    \caption{
    \textbf{Cumulative ablation study} for the proposed modules and losses of RGNet. The experiment is conducted w/o NaQ augmentation on the Ego4D dataset. In each step, we remove one of our proposed modules or losses to evaluate their individual impact. $^{\ast}$denotes the baseline model.
  }
   \vspace{-2em}
  \label{tab:ablation_module}
\end{table}

\hspace{1pt} \\
\textbf{Modules:}
In our initial experiments, we validate the effectiveness of each proposed module with a cumulative ablation in Tab. \ref{tab:ablation_module}.
The unified network demonstrates superior performance across all evaluation metrics, notably achieving an R$1_{.3}$ score of $18.28\%$. Note that the disjoint baseline performs $4.2\%$ worse than our unified network. This performance difference can be attributed to our models unified architecture. 

With the removal of the \textbf{retrieval token} for generating the clip feature in Eq. \ref{eq:fuse}, considerably drops the score by about $0.5\%$. This proves that modeling both clip level context and frame level content are necessary to produces strong multimodal features for the grounding decoder. Subsequently, eliminating the \textbf{sparsifier}, there is a decline in R$1_{.3}$ performance of approximately $1.7\%$. This verifies the impact of the sparsifier in modeling the fine-grained event boundaries. Finally, removing the \textbf{RG-Encoder} yields a notable degradation of approximately $2.0\%$ in the R$1_{.3}$ score, which substantiates its capability to model clip and frame level granularity jointly.

\hspace{1pt} \\
\textbf{Loss Functions:} 
Training RGNet without our \textbf{negative clip sampling} strategy for the contrastive loss (Tab. \ref{tab:ablation_module}) results in an almost $0.9\%$ decline in the R$1_{.3}$ score. Our sampling strategy mimics long-video setup closely during training, which improves event discrimination capability inside the same scene and surroundings. 

By eliminating the \textbf{attention loss}, we observe a further reduction of $1.2\%$ in grounding performance. The attention loss enables our encoder to model fine-grained events in long videos which is crucial for specific moment detection. 

\begin{figure}[!h]
    \centering
    \begin{tabular}{c|c}
      \includegraphics[width=.3\linewidth]{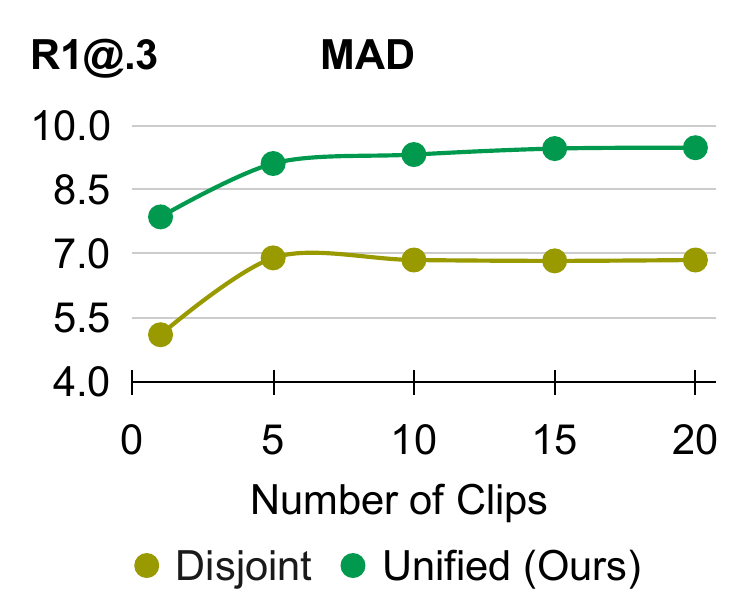}   &
      \includegraphics[width=.3\linewidth]{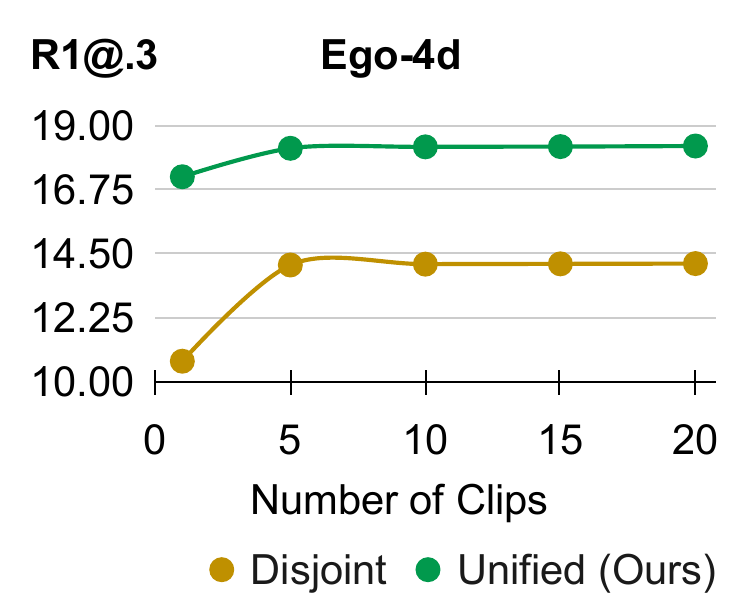}
    \end{tabular}
    \caption{\textbf{Impact of number of retrieved clips.} Reducing the number of clips speeds up the network execution time. While the baseline model experiences a significant drop in performance with this reduction, RGNet shows a noticeably smaller decline in performance under the same conditions.}
    \vspace{-2em}
    \label{fig:proposal_number}
\end{figure}

\hspace{1pt} \\
\textbf{Number of Clips:} 
The number of retrieved proposal clips significantly influences the Long Video Temporal Grounding (LVTG) performance. Our model consistently outperforms the baseline across various choices of top-k retrieved clips, as shown in Fig. \ref{fig:proposal_number}. Even with only a single retrieved clip, our performance surpasses the baseline's optimal result for both datasets. Moreover, our model exhibits a less steep performance decline when fewer clips are retrieved. Importantly, with fewer clips, the grounding network runs fewer times, leading to an overall speedup. Therefore, our model strikes a favorable balance between efficiency and performance, showcasing superior performance than the baseline.

\begin{figure}[!h]
    \centering
    \begin{tabular}{cc}
      \includegraphics[width=.3\linewidth]{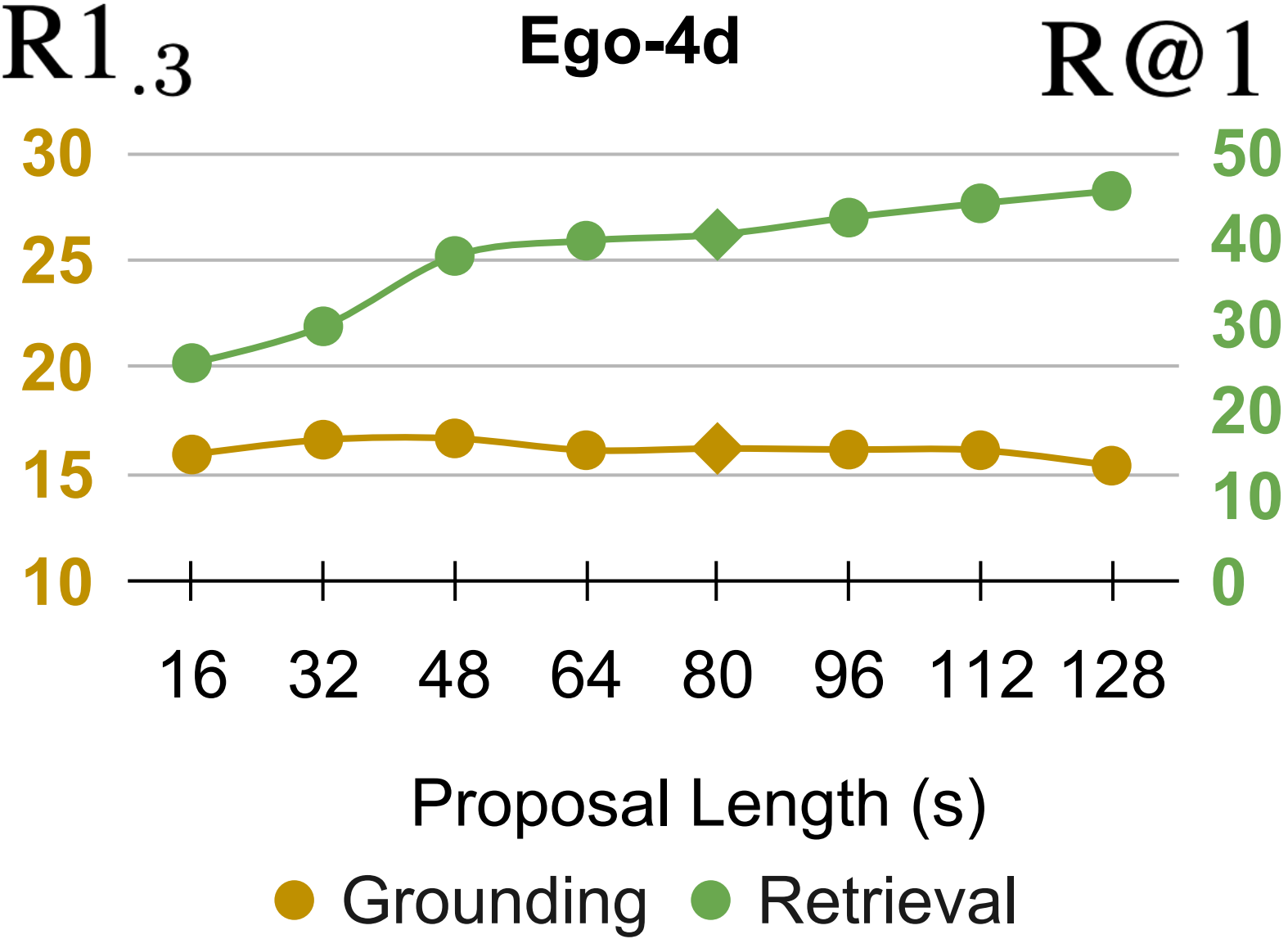}   &
      \includegraphics[width=.3\linewidth]{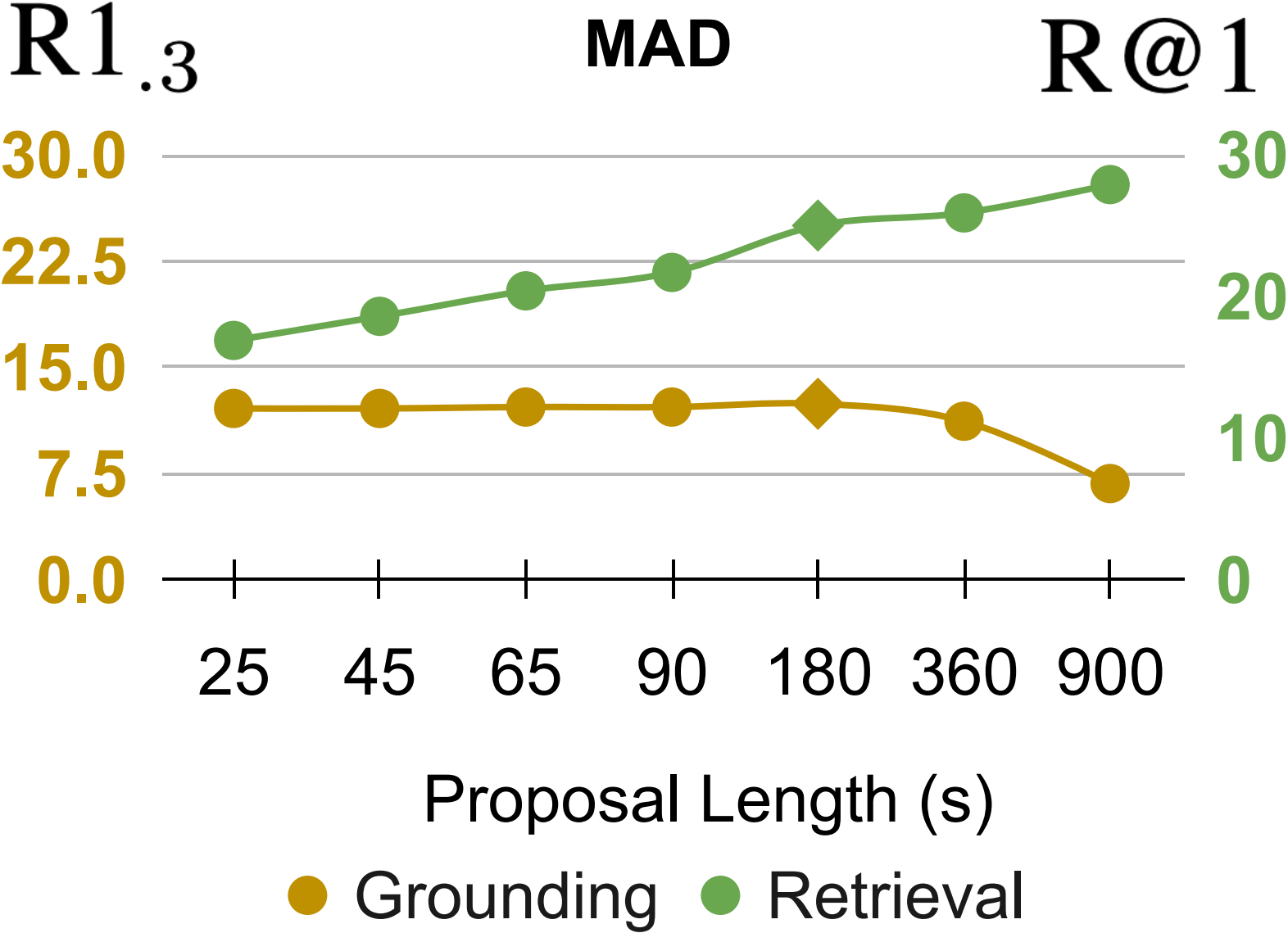}
    \end{tabular}
    \caption{\textbf{Impact of retrieved clip length.} Longer clips result in improved retrieval due to fewer candidates. However, grounding becomes exceedingly difficult with longer clips. For example, the grounding performance drops after 180 seconds and 48 seconds for the MAD and Ego4D datasets.}
    \vspace{-2em}
    \label{fig:proposal_length}
\end{figure}

\hspace{1pt} \\
\textbf{Clip Length:} 
Varying clip length impacts both retrieval and grounding performance. The number of clip decreases with longer clips, making the retrieval task easier. Fig. \ref{fig:proposal_length} reports a monotonically improving retrieval with longer clips.  However, after a certain length, moment localization quality starts to deteriorate. With longer clips, the task becomes extremely difficult. We see the decline after the 180s and 48s for the MAD and Ego4D datasets. Hence, we set the default clip length of our method to these optimal lengths. Importantly, the unified network helps outscore the disjoint baseline on all tested clip lengths (inversely proportional to the clip numbers in Fig. \ref{fig:proposal_number}). 

\subsection{Qualitative Analysis}
\begin{figure*}[!h]
    \centering
    \includegraphics[width=.8\linewidth]{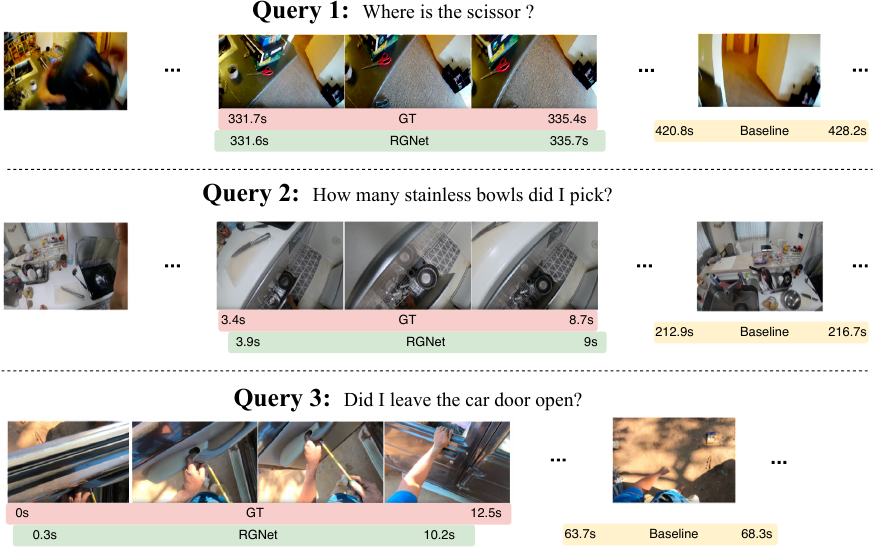}
    \caption{\textbf{Qualitative Results.} The baseline fails to retrieve the correct clip in the first two queries. Since they primarily depict the same indoor room throughout the whole video, precise event discrimination is vital for accurate clip retrieval. In the third query, the baseline cannot identify the moment within the correctly retrieved clip, detecting it only after it has concluded. Precise event localization demands improved alignment between visual events and the queried text. RGNet correctly localizes all the moments.}
    \vspace{-3.5em}
    \label{fig:ego_vis}
\end{figure*}

We visualize the qualitative predictions of RGNet compared to the disjoint baseline in Fig. \ref{fig:ego_vis}. For the first and second queries, the baseline struggles to retrieve the correct clip where the scissor appears. RGNet accurately retrieves the clip with the searched object and precisely localizes the moment it appeared in the video. Disjoint retrieval often leads to incorrect clip selection, which the final grounding network cannot recover from. For the third query, the baseline selects the correct clip but fails to localize the precise moment the man closes the car door, indicating a gap between the two stages. RGNet mitigates this gap by leveraging features from the retrieved clip to improve moment localization. More visual outputs are included the supplementary materials.

\section{Conclusion}
We introduce an end-to-end model for long video temporal grounding, unifying the two stages of prevailing methods with shared features and mutual optimization. We conduct independent analyses of the two stages, shaping our solution based on their distinct impacts. This leads to a better understanding of fine-grained events in long videos. Our approach demonstrates state-of-the-art performance on challenging long video grounding datasets, validating its effectiveness. Like all LVTG methods, we rely on a pre-trained image encoder to extract the visual features. A promising future direction includes eliminating the need for an image encoder and training directly on the raw video data.


%
%
\bibliographystyle{splncs04}
\bibliography{main}
\clearpage
\appendix
\section{Supplementary} 
\subsection{Further Technical Deatails}
\noindent \textbf{Motivation of contextual feature} $\mathbf{R}$. Most prior works use a heuristic-based scoring function borrowed from the short video-text retrieval methods and apply them directly to longer videos. This often leads to poor performance in fine-grained event retrieval within very long videos. Instead, we use a learnable retrieval token \textbf{R}, facilitating more effective fine-grained event understanding in long videos. Our introduced retrieval token unifies the retrieval and grounding objectives since it is used as input into the grounding decoder. This enables our model to maximize the synergy between retrieval and grounding stages, significantly improving both clip retrieval (\textbf{+12.6\% R1}) and grounding (\textbf{+4.2\% R1@.3}) on Ego4D (Tab. \ref{tab:r}).
For both training and inference, we randomly initialize $\mathbf{R}$ and then replicate it for all clips. Thus, the initial features $\mathbf{R}_i$ and $\mathbf{R}_j$ of clips $i$ and $j$ are replicated copies of a randomly initialized feature $\mathbf{R}$. 
\vspace{-2em}
\begin{table}[h]
    \centering
    \footnotesize
    \begin{tabular}{cccrcc}
            \multirow{2}{*}{Method} & \multicolumn{2}{c}{Grounding} & &
    \multicolumn{2}{c}{Retrieval} \\
         \cmidrule{2-3}\cmidrule{5-6} & R1@.3 ($\uparrow$)  & R5@.3 ($\uparrow$)  && R1 ($\uparrow$)  & R5 ($\uparrow$) \\
         \hline
         Heuristic & 14.15 & 30.33 && 12.41 & 24.50 \\
         \textbf{Learned} & \textbf{18.28} & \textbf{34.02} && \textbf{25.01} & \textbf{50.02} \\
        \hline
    \end{tabular}
    \caption{Performance Comparison on Ego4D (ref. Tab 2, 3)}
    \label{tab:r}
    \vspace{-3em}
\end{table}

\noindent \textbf{Runtime Analysis.} We divide a video of length $L$ into clips of length $C$. The number of text tokens is $T$, and the hidden dimension is $D$. The time complexity of self-attention is $O(LCD)$, cross-attention is $O(LTD)$, and the full model is $O(L(C+T)D)$. The runtime of our model scales linearly with video length. We compared our runtime with the CONE baseline on a GTX A100 and reported the total time for the Ego4D validation set. CONE achieves a $14.15\%$ R1@0.3 score with a 39.9-second inference time. In contrast, RGNet scores $20.63\%$ R1@0.3 in just 24.2 seconds, making it \textbf{1.7x} faster than CONE. Our retrieval module achieves superior performance with fewer retrieved clips, substantially reducing the runtime, as shown in the leftmost subfigure below.

\begin{figure}[!h]
    \centering
    \begin{tabular}{cc}
    \includegraphics[width=.45\linewidth]{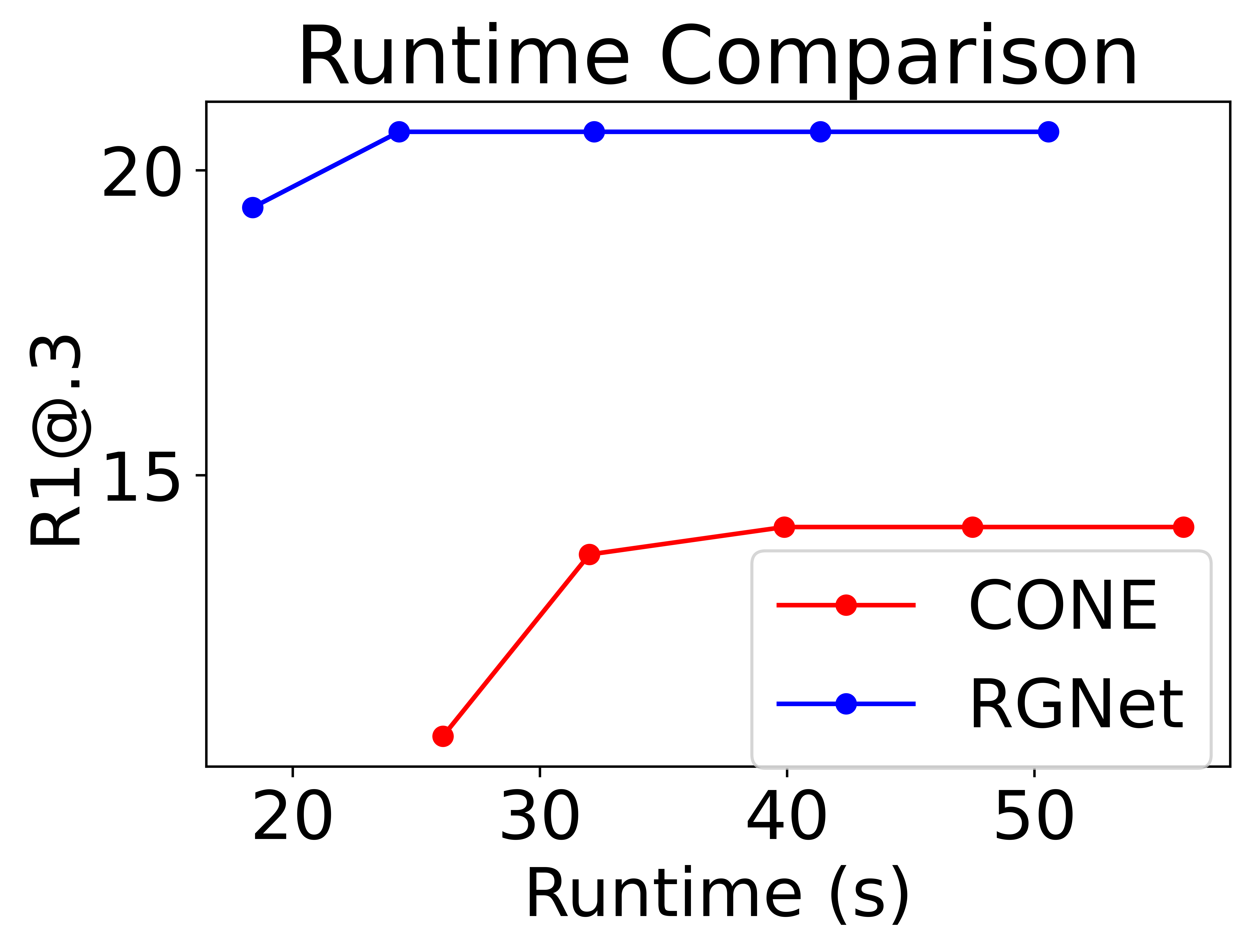}    & 
    \includegraphics[width=.45\linewidth]{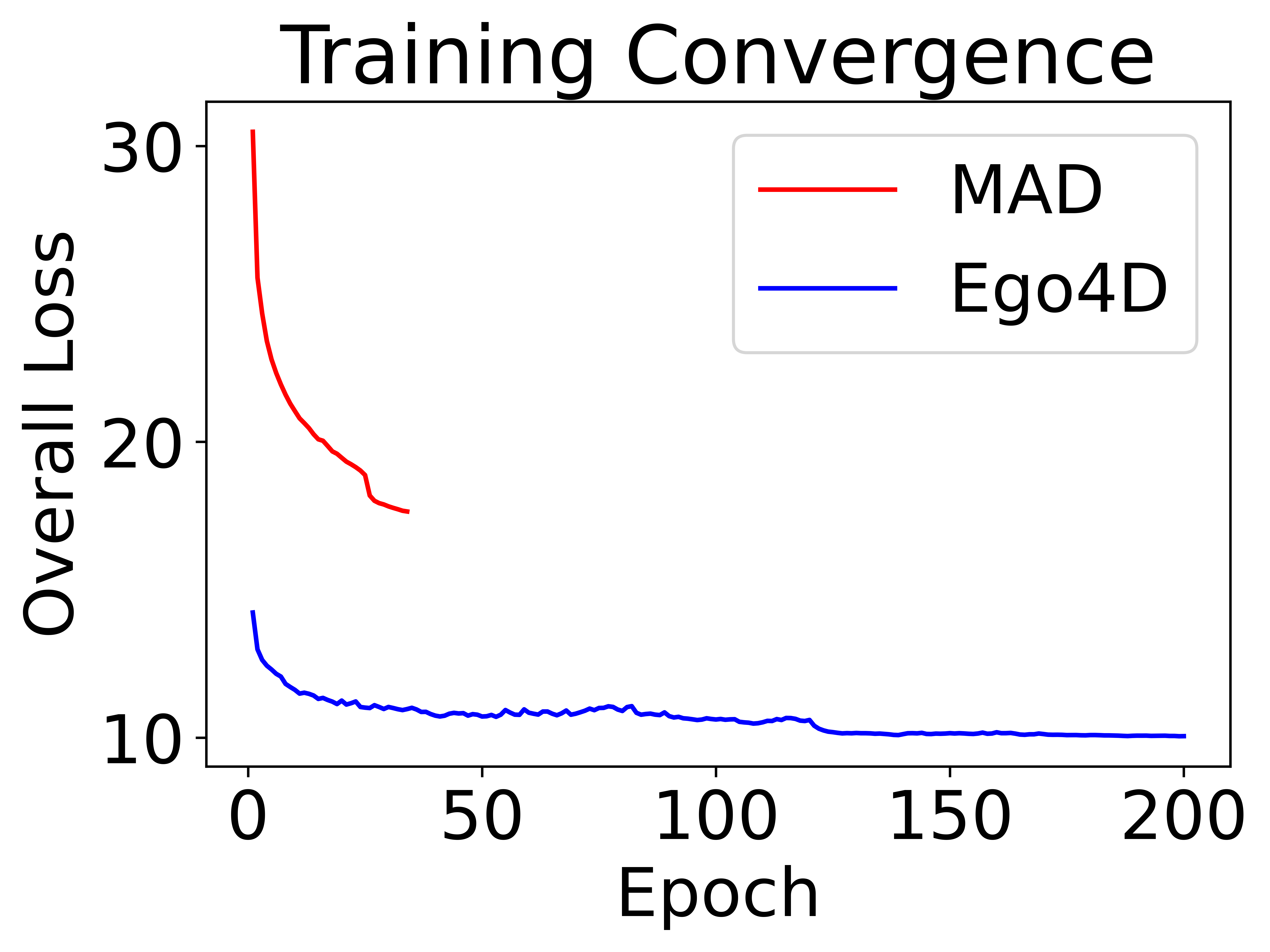}
    \\
    \end{tabular}
    \label{tab:my_label}
    \caption{\textbf{(left)} Runtime comparison w.r.t. baseline model CONE. \textbf{(right)} Training convergence curve for both MAD and Ego4D datasets.}
\end{figure}

\noindent \textbf{Convergence.} RGNet converges in 35 and 200 epochs on MAD and Ego4D. The rightmost figure above shows that on MAD our model converges earlier due to a larger dataset.

\noindent \textbf{Number of the Negative Samples.} We trained Ego4D with a batch size of 32, resulting in 1024 clip-text pairs (32 positives and 992 negatives). We use 8 positive and 56 negative samples for MAD to fit its longer clips in a GPU.

\subsection{Further Ablation Studies} This section provides an in-depth analysis of various components of RGNet, such as the sparsifier and transformer decoder. Furthermore, we conduct ablation studies on pre-training data and frame rate. All experiments are carried out on the Ego4D-NLQ dataset.

\hspace{1pt}\\
\textbf{Sparsifier:}
We use a differentiable softmax function to enable end-to-end training of the sparsifier. Let's denote $p^j$ as the linear projection of $f^j$ from Eq. \ref{eqn:sampler}.
\begin{equation}
    p^j = \text{linear}({f}^{j})
\end{equation}
To derive a categorical variable $G^{j}$ characterized by probabilities ${\pi_1}^{j} = \sigma(p^j)$ and ${\pi_0}^{j} = 1- \sigma(p^j)$, where $\sigma$ is the sigmoid operation, we can reframe the sampling procedure for $G^{j}$ through the utilization of the Gumbel-Max trick, outlined as follows:
\begin{equation}
G^{j}={\arg \max}_k \left\{\log \left({\pi}_{k}^j\right)+g_k:k=0,1\right\}
\end{equation}
Here, the set $\{g_k\}_{k={0,1}}$ consists of independently and identically distributed (i.i.d.) random variables sampled from the $Gumbel(0,1)$ distribution. Considering the non-differentiable characteristic of the argmax operation, we employ an approximation for $G^{j}$ using a differentiable, soft version $\hat{G}^{j}$, derived from the Gumbel-Softmax relaxation  \cite{gumble-softmax, gumble-softmax_2}.

\begin{equation}
\hat{G}^{j}=\frac{\exp \left(\left(\log \left({\pi}_1^j\right)+g_1\right)/{\tau}\right)}{\Sigma_{k \in{0,1}} \exp \left(\left(\log \left({\pi}_k^j\right)+g_k\right)/\tau\right)}
\label{eqn:gumbel-softmax}
\end{equation}

To ensure differentiability with respect to the discrete samples \(G^j\), we employ the straight-through trick~\cite{gumble-softmax} and utilize the gradients of \(\hat{G}^j\) as an approximation for the gradients of \(G^j\) in the backward pass. 

\begin{figure*}
     \centering
     \begin{subfigure}[b]{0.3\textwidth}
         \centering
\includegraphics[width=1\linewidth]{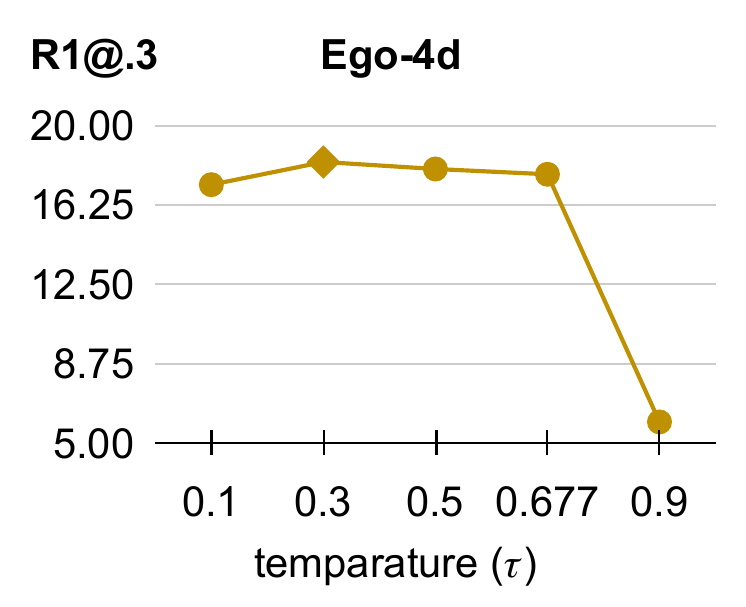}
         \caption{Low temperature enables our sparsifier to approximate a categorical distribution over the relevancy of frames and achieves better performance.}
         \label{fig:eps}
     \end{subfigure}
     \hfill
     \begin{subfigure}[b]{0.3\textwidth}
         \centering
         \includegraphics[width=1\linewidth]{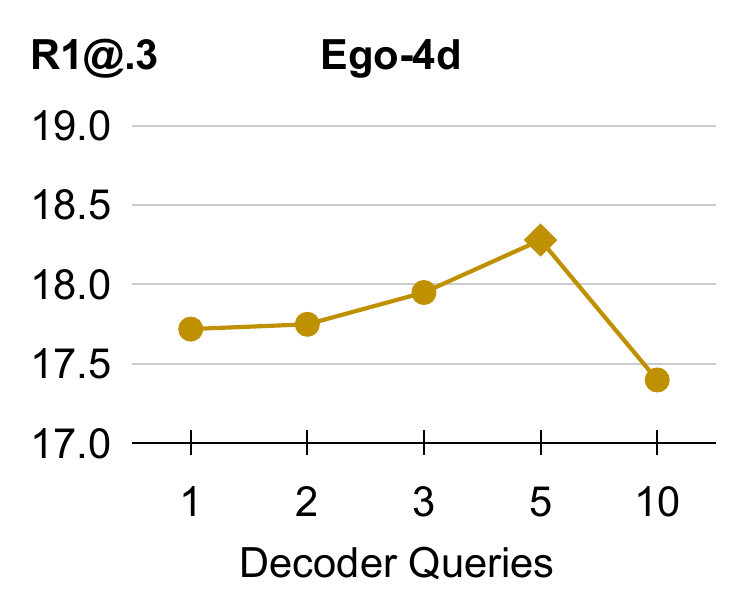}
         \caption{Increasing the number of decoder queries up to $5$ enhances performance, while further increments start to deteriorate the results.}
         \label{fig:decoder}
     \end{subfigure}
     \hfill
     \begin{subfigure}[b]{0.3\textwidth}
         \centering
\includegraphics[width=1\linewidth]{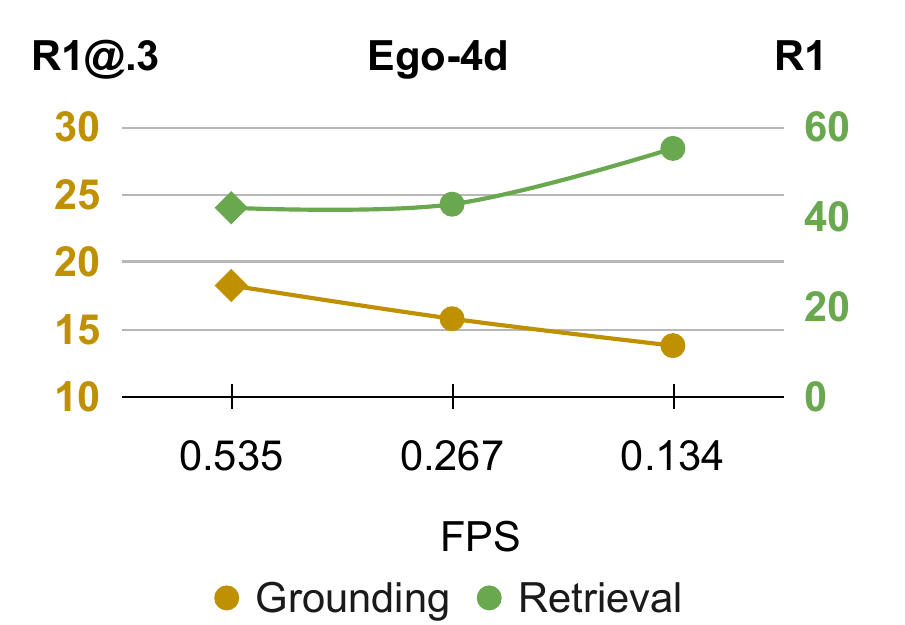}
         \caption{While decreasing the FPS improves retrieval accuracy, it deteriorates the grounding performance due to the loss of temporal information.}
        \label{fig:fps}
     \end{subfigure}
        \caption{\textbf{Ablation Studies} on the sparsifier, transformer decoder, and input FPS.}
        \label{fig:three graphs}
\vspace{-3em}
\end{figure*}

The Gumbel-Softmax distribution serves as an interpolation between discrete one-hot-encoded categorical distributions and continuous categorical densities. For low temperatures (\(\tau\) = 0.3), the expected value of a Gumbel-Softmax random variable approaches the expected value of a categorical random variable with the same logits. As the temperature increases (\(\tau\) = 0.9), the expected value converges to a uniform distribution over the categories. Fig. \ref{fig:eps} shows that we achieve the best performance with \(\tau\) to be \(0.3\), which means the differentiation between the relevant and irrelevant frames is beneficial during the retrieval.

\hspace{1pt}\\
\textbf{Transformer Decoder:} RGNet utilizes learnable decoder queries \cite{dabdetr} to localize the moments. The number of queries equates to the number of predicted moments from each retrieved proposal. With more queries, the decoder can detect moments in different temporal locations of various widths. However, for long videos, the decoder runs on multiple retrieved clips, and increasing the queries beyond $5$ leads to an increasing number of predicted moments, which decreases the LVTG performance (ref. Fig. \ref{fig:decoder}). Consequently, we set the number of queries in our decoder to $5$. 

\hspace{1pt}\\
\textbf{Frame Rate:}
Manual reduction of frame rate (FPS) yields enhanced retrieval accuracy but significantly degrades grounding performance, as shown in Fig. \ref{fig:fps}. Lowering the FPS serves as a heuristic sparsification strategy that aids retrieval but results in temporal information loss, leading to diminished overall performance. In contrast, our learned sparsifier enhances retrieval accuracy without compromising grounding performance, presenting a superior alternative to manual sparsification.
\begin{table}[!h]
\centering
\resizebox{.5\linewidth}{!}{
\begin{tabular}{lccccc}
\toprule 
Method  & \naq &  R1$_{.3}$ & R5$_{.3}$ &R1$_{.5}$ & R5$_{.5}$       \\
  \midrule
VSLNet &  \xmark   &      5.45      &     10.74      &    3.12      &    6.63      \\
EgoVLP  &  \xmark   &      10.84     &     18.84      &    6.81      &    13.45     \\
ReLER &  \xmark   &      14.66     &     17.84      &    8.67      &    11.54     \\ 
RGNet (Ours) & \xmark & \textbf{18.28} & \textbf{34.02} & \textbf{12.04}  & \textbf{22.89} \\ 
 
\midrule

VSLNet &  \cmark   &   {10.26}   & {19.01}     &{5.81}     &{12.67}    \\ 
EgoVLP  &  \cmark   &   {15.90}   & {26.38}     &{9.46}     &{17.80}    \\ 
ReLER &  \cmark   &   {19.31}   & {23.62}     &{11.59}    &{15.75}    \\
RGNet (Ours) &  \cmark & \textbf{20.63} & \textbf{41.67} &\textbf{12.47} &\textbf{25.08} \\

\bottomrule
\end{tabular}
}
\caption{\textbf{Impact of NaQ.} We compare RGNet on the Ego4D-NLQ dataset with and w/o NaQ annotations. RGNet achieves the best performance in both cases.}
\label{tab:naq-results}
\vspace{-3em}
\end{table}

\hspace{1pt}\\
\textbf{NaQ Pretraining:}
Similar to the prior state-of-the-art model \cite{ramakrishnan2023naq}, we employ NaQ annotations to pre-train RGNet on the Ego4D dataset. The grounding annotations of NaQ are automatically generated from the ground truth narrations of the Ego4D dataset. With this pretraining, we improve R$@1_{.3}$ and R$@1_{.5}$ by \textbf{1.32\%} and \textbf{18.05\%} (refer to Table \ref{tab:naq-results}). Importantly, without the extra NaQ annotations, RGNet demonstrates a larger improvement of \textbf{3.62\%} for R$@1_{.3}$. These results highlight RGNet's superior performance, even in scenarios with limited data. The evaluation of the Ego4D test set is exclusively available on their official server, which is presently closed. Therefore, in alignment with recent publications \cite{zhang2023helping, pan2023scanning} in ICCV'23, we present our performance of Ego4D-NLQ on the validation set.

\subsection{Qualitative Results}
\begin{figure*}
    \centering
    \includegraphics[width=1\linewidth]{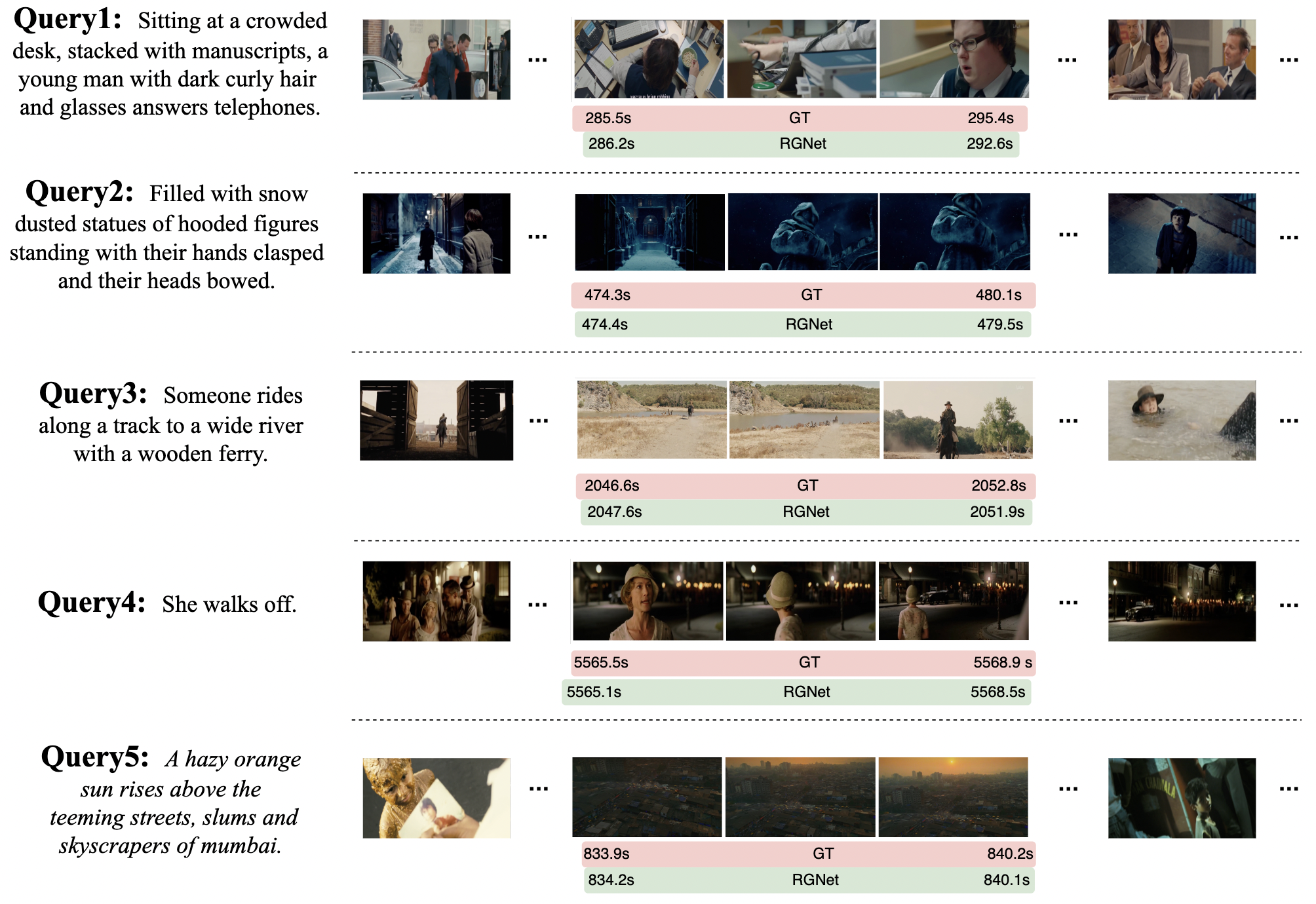}
    \caption{\textbf{Qualitative results on MAD.} RGNet successfully localizes moments from hour-long movies by parallelly processing them in clip and frame level granularity. }
    \label{fig:mad_supp}
\end{figure*}

\begin{figure*}
    \centering
    \includegraphics[width=1\linewidth]{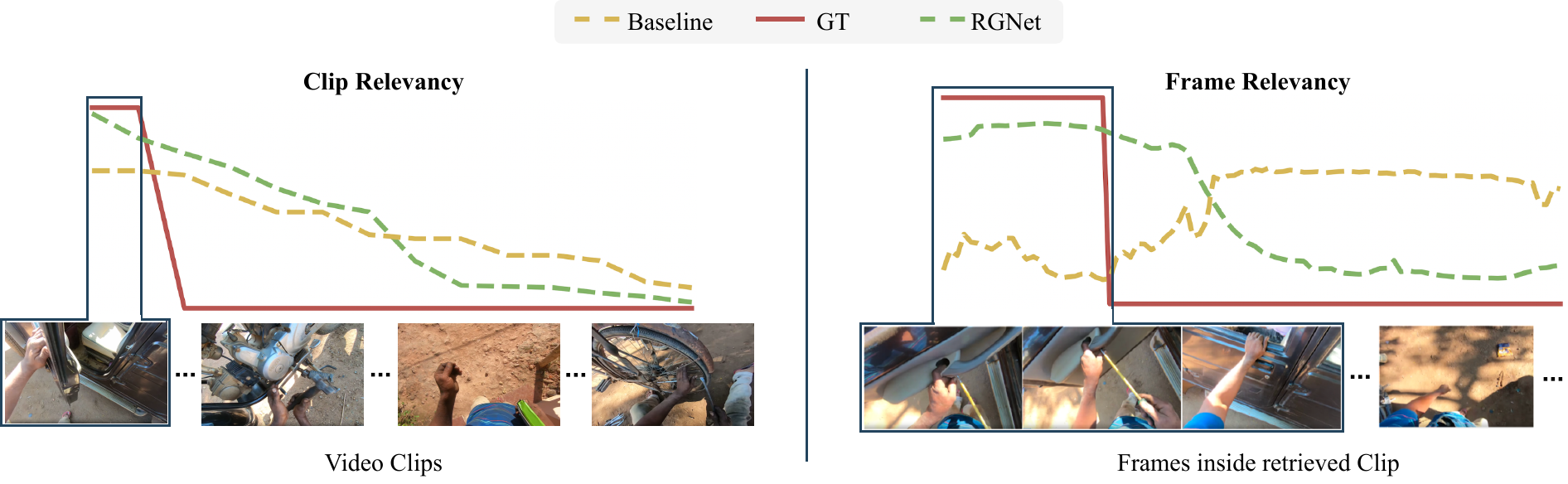}
    \caption{\textbf{Clip and Frame Relevancy.} We visualize the relevancy score of video clips \textbf{(left)} and proposal frames \textbf{(right)} for the query \quotes{\textit{Did I leave the car door open}?} from Ego4D-NLQ. We present the score for both RGNet and the disjoint baseline model in green and yellow, respectively. RGNet approximates the ground truth clip and frames better than the baseline in both stages. }
    \label{fig:relevancy}
\end{figure*}

\begin{figure*}
    \centering
    \includegraphics[width=1\linewidth]{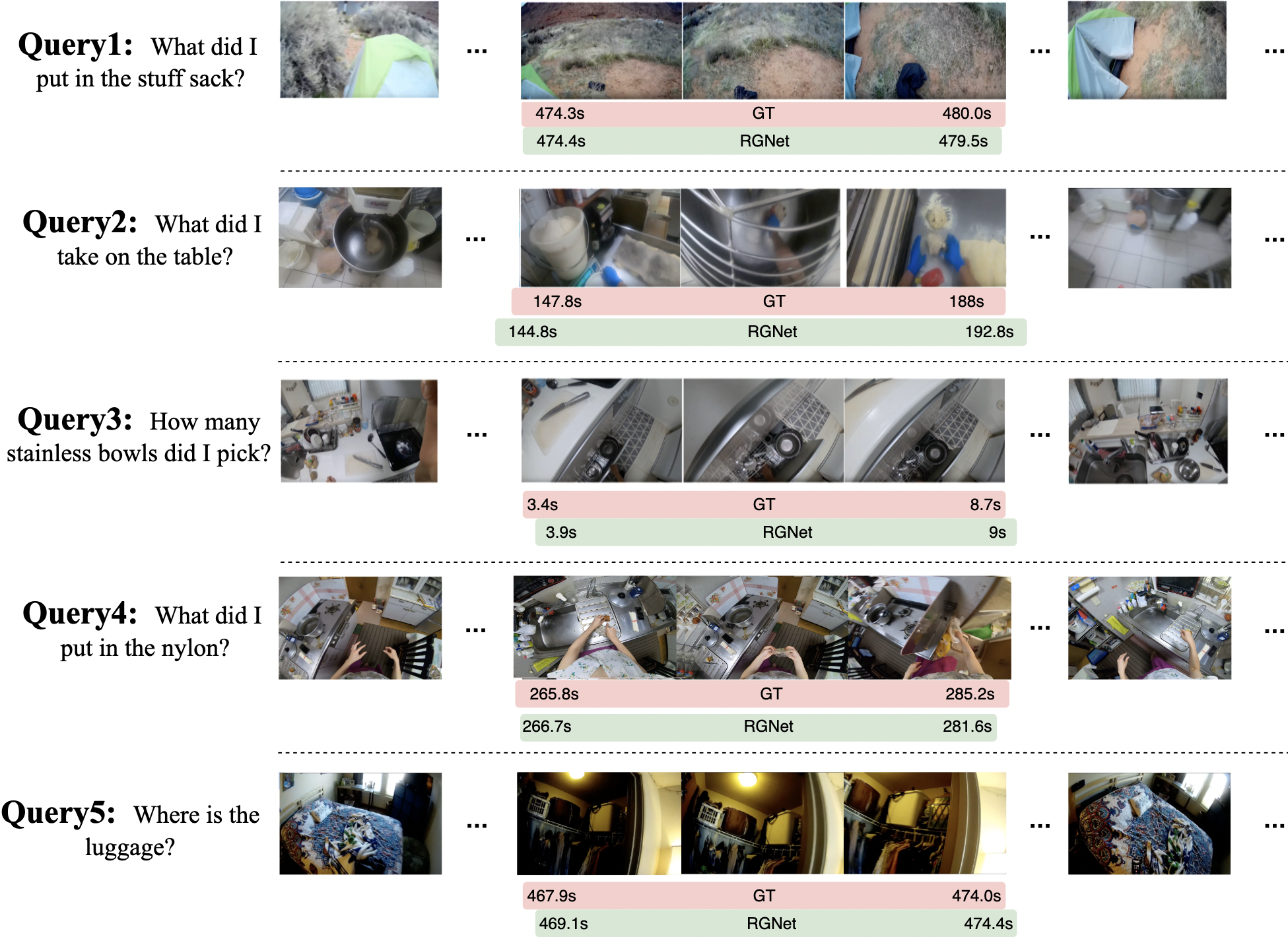}
    \caption{\textbf{Qualitative results of Ego4D.} RGNet can localize fine-grained events in long videos across various scenes and scenarios. }
    \label{fig:ego_supp}
\end{figure*}

We visualize the relevance score of all the video clips and clip frames for a single video in Fig. \ref{fig:relevancy}. For both the clip and frame levels, our model scores higher on the ground truth regions than the baseline. Then, we visualize some successful moment localization on both MAD and Ego4D datasets in Tab. \ref{fig:mad_supp} and \ref{fig:ego_supp}.

\end{document}